\documentclass[journal]{IEEEtran} 
\IEEEoverridecommandlockouts

\usepackage[pdftex]{graphicx}
\usepackage{subcaption}
\usepackage{booktabs} 

\usepackage{amsmath}

\def\BibTeX{{\rm B\kern-.05em{\sc i\kern-.025em b}\kern-.08em
    T\kern-.1667em\lower.7ex\hbox{E}\kern-.125emX}}

\begin{document}

\title{Addressing Annotation Scarcity in Hyperspectral Brain Image Segmentation with Unsupervised Domain Adaptation}

\author{
    \IEEEauthorblockN{
        Tim Mach\textsuperscript{1},
        Daniel Rueckert\textsuperscript{1,2,3},
        Alex Berger\textsuperscript{1},
        Laurin Lux\textsuperscript{1,3}
        Ivan Ezhov\textsuperscript{1},
    }
    \\
    \IEEEauthorblockA{\textsuperscript{1}\textit{Technical University of Munich, Munich, Germany}}
    \IEEEauthorblockA{\textsuperscript{2}\textit{Department of Computing, Imperial College London, London, UK}}
    \IEEEauthorblockA{\textsuperscript{3}\textit{Munich Center for Machine Learning (MCML), Munich, Germany}},
    \IEEEauthorblockA{tim.mach@tum.de}
}

\maketitle

\begin{abstract}
This work presents a novel deep learning framework for segmenting cerebral vasculature in hyperspectral brain images. We address the critical challenge of severe label scarcity, which impedes conventional supervised training. Our approach utilizes a novel unsupervised domain adaptation methodology, using a small, expert-annotated ground truth alongside unlabeled data. Quantitative and qualitative evaluations confirm that our method significantly outperforms existing state-of-the-art approaches, demonstrating the efficacy of domain adaptation for label-scarce biomedical imaging tasks.
\end{abstract}

\section{Introduction}
Accurate delineation of cerebral blood vessels is paramount during neurosurgical procedures to minimize the risk of iatrogenic injuries, such as haemorrhaging, which can cause severe neurological deficits. Intraoperative imaging techniques are crucial for providing surgeons with real-time guidance to navigate the complex brain anatomy. Hyperspectral Imaging (HSI) has emerged as a promising non-invasive modality in this context. By capturing a wide spectrum of light beyond the visible range, HSI provides rich spectral signatures that can distinguish between different tissue types, such as arteries, veins, and brain parenchyma, with high fidelity \cite{HSIclinical}. Automating the segmentation of the vascular network from these HSI cubes can offer an objective, reliable map to guide surgical interventions, enhancing safety and improving patient outcomes.

Deep learning, particularly convolutional neural networks (CNNs) \cite{unet}, has become the state-of-the-art for medical image segmentation tasks due to its ability to learn complex hierarchical features. However, the performance of these models is heavily reliant on the availability of large, meticulously annotated datasets for training. This presents a significant bottleneck in the medical domain, where expert annotation is a costly and time-consuming process. The HELICoiD dataset, while being a valuable resource for brain HSI analysis, exemplifies this challenge with its limited number of labeled samples. This scarcity of data hinders the effective training of conventional segmentation architectures and motivates the exploration of data-efficient learning strategies.

Transfer learning represents a compelling direction to mitigate data scarcity. However, from our initial attempts, traditional transfer learning methods, despite resulting in high precision in pixel-level vessel prediction, do not address the continuity of the entire blood vessel tree. Building on these insights, the present work aims to predict the entire vessel tree, ensuring structural continuity.
To achieve this, we investigate advanced deep learning methodologies, including domain adaptation and dimensionality reduction methods. We systematically compare these approaches to identify the most effective methods for robust blood vessel segmentation in hyperspectral images.

\section{Methods}
\subsection{Model Architecture}
\subsubsection*{Unsupervised Domain Adaptation}
\begin{figure}[h]
\centering
\includegraphics[width=\columnwidth]{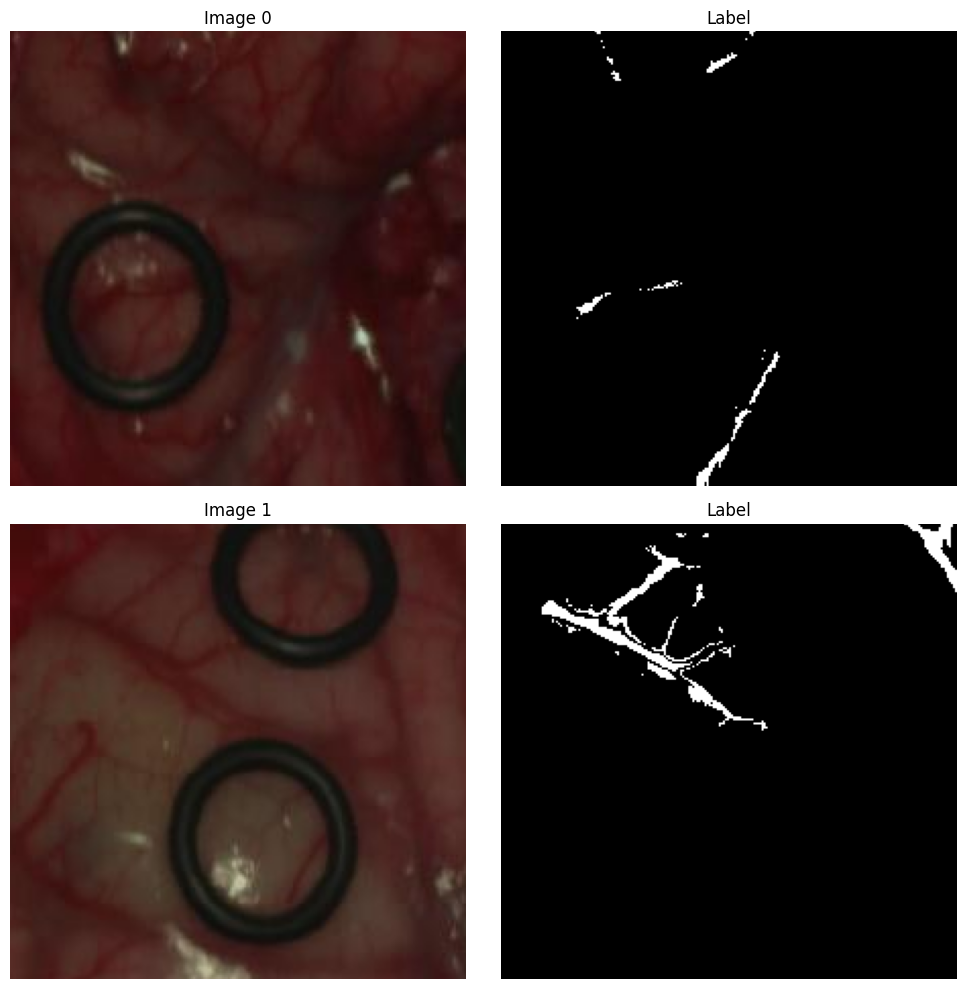}
\caption{Example data from HELICoid, where the left column is an RGB image and the right column is the labels of the blood vessels.}
\label{fig:helicoid}
\end{figure}

\begin{figure}[t]
\centering
\includegraphics[width=\columnwidth]{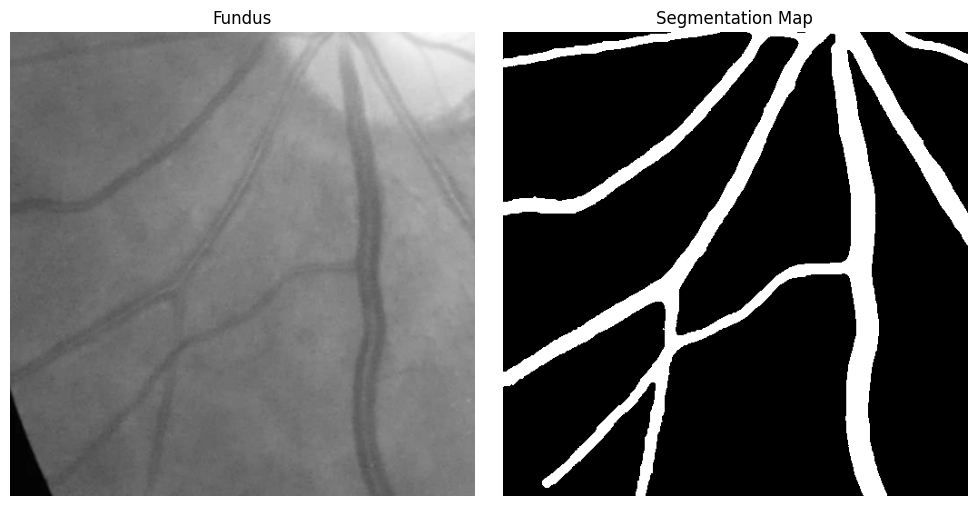}
\caption{Example data from FIVES, where the left column is a grayscale image and the right column is the segmentation map of the blood vessels.}
\label{fig:random_crop}
\end{figure}

Our approach leverages unsupervised domain adaptation to bridge the domain gap between the source and target domains. We pre-train a segmentation network on the source dataset (FIVES  \cite{fives}, Fig. \ref{fig:random_crop}), which contains fundus images annotated at the pixel level for blood vessel structures, and test on the target dataset (HELICoiD \cite{helicoid}, Fig. \ref{fig:helicoid}).

\begin{figure*}[t]
    \centering
    \begin{subfigure}[b]{0.49\textwidth}
        \centering
        \includegraphics[width=\textwidth]{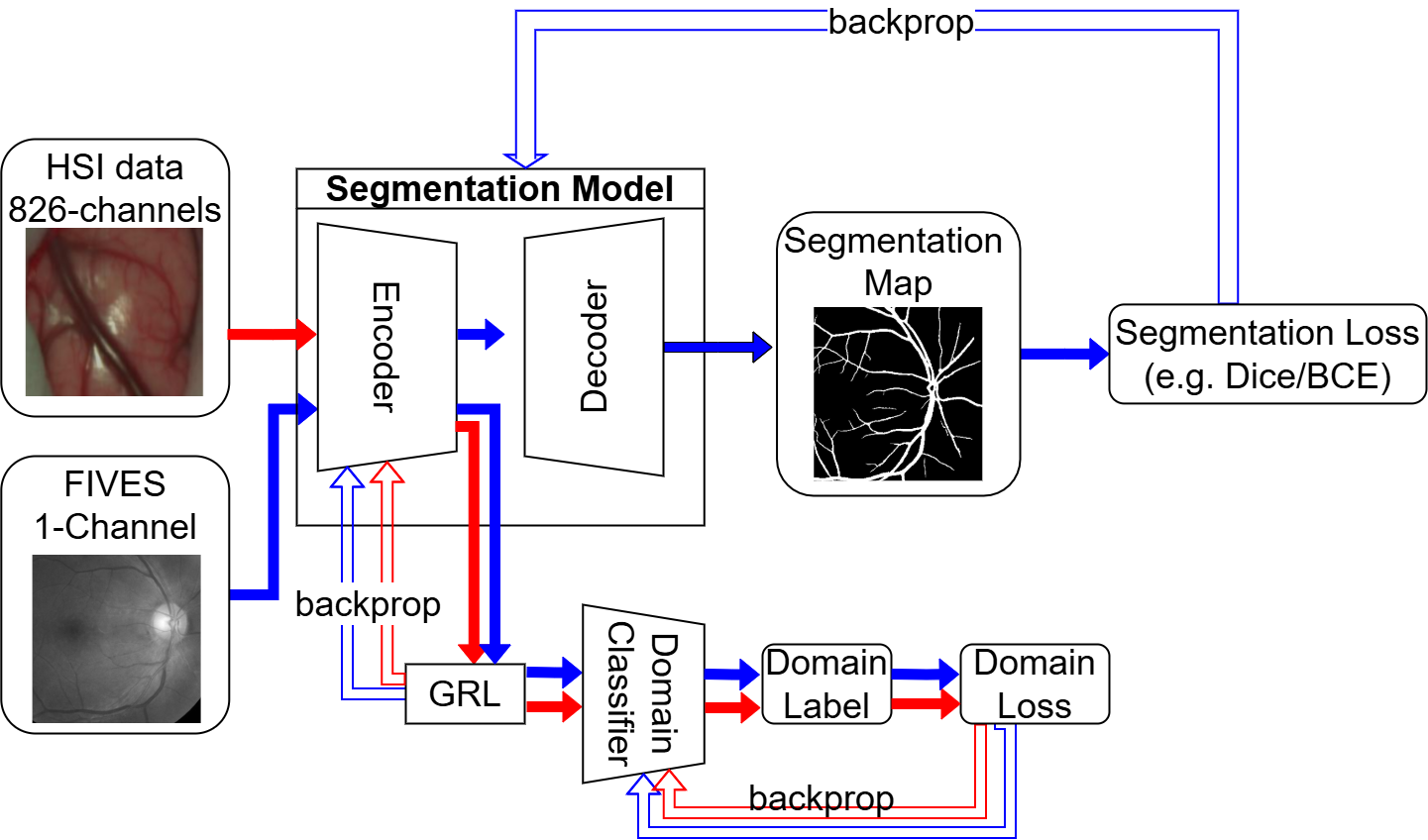}
        \caption{Domain Adaptation architecture tailored to our datasets and blood vessel segmentation.}
        \label{fig:domain_adaptation}
    \end{subfigure}
    \hfill
    \begin{subfigure}[b]{0.49\textwidth}
        \centering
        \includegraphics[width=\textwidth]{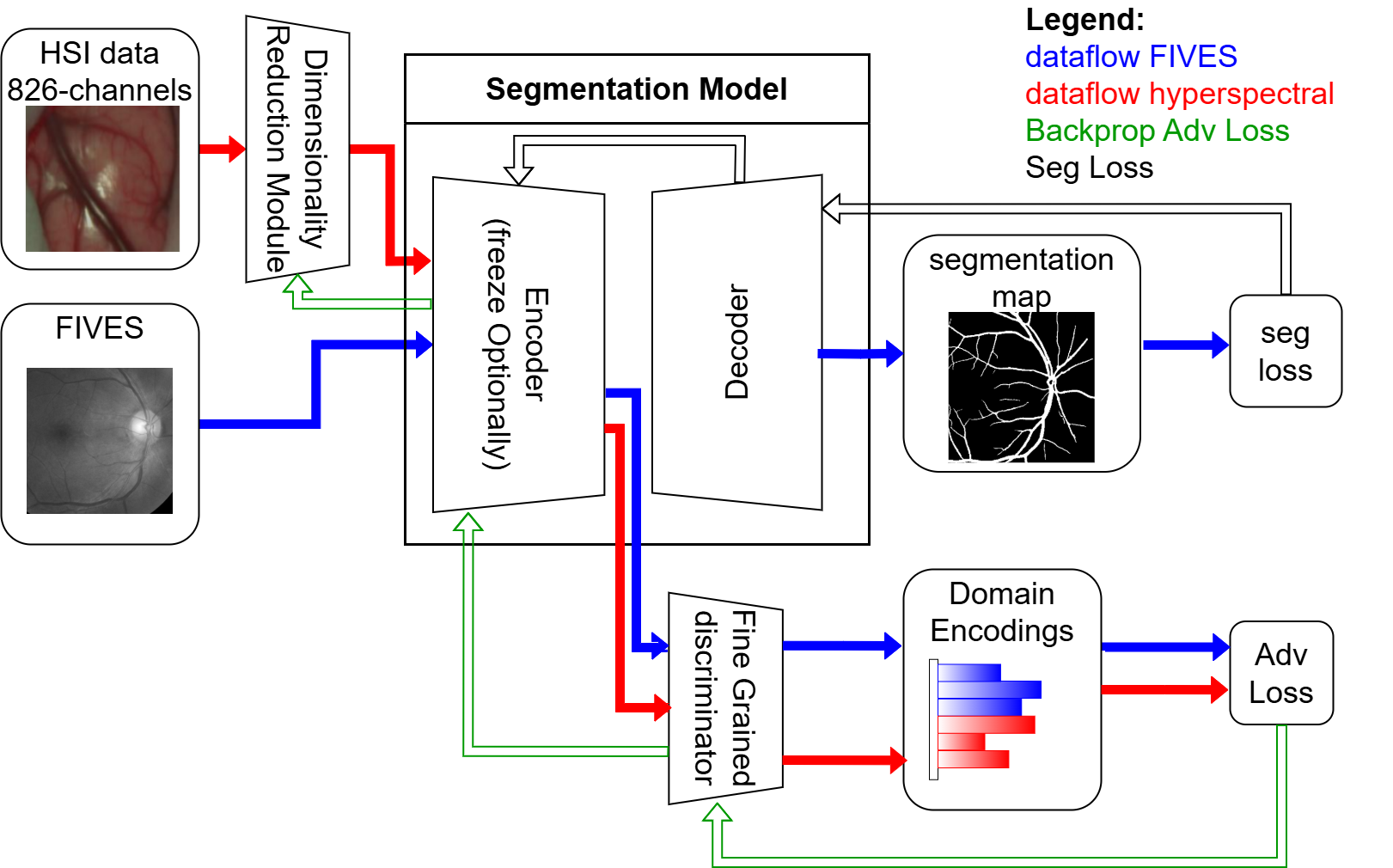}
        \caption{FADA architecture tailored to our datasets and blood vessel segmentation.}
        \label{fig:fada}
    \end{subfigure}
    \caption{Architectures for domain adaptation and feature alignment.}
    \label{fig:domain_adaptation_architecture}
\end{figure*}

Building on the framework proposed in \cite{DA_GRL}, we adapt the conventional domain adaptation architecture, originally devised for classification tasks, to the segmentation problem. As illustrated in Figure \ref{fig:domain_adaptation_architecture}a, our architecture comprises an encoder-decoder segmentation network augmented with a domain classifier branch. The appended domain classifier receives the encoder’s feature representations and predicts whether they originate from the source or target domain.

Training is performed in a dual-optimization manner. The segmentation branch is supervised using a standard segmentation loss computed on the source domain images. Concurrently, the domain classifier is trained using a domain adversarial loss that is backpropagated through both the classifier and the encoder. This adversarial mechanism compels the encoder to learn domain-invariant features, thus adapting the segmentation model to the HELICoiD dataset.

\subsubsection*{Fine-grained Adversarial Approach}
In the conventional unsupervised domain adaptation framework \cite{DA_GRL}, the domain classifier distinguishes whether a feature vector originates from the source or target domain. However, this global adversarial loss is often insufficient for pixel-wise segmentation tasks, as it does not capture the class-specific discrepancies inherent in segmentation. To address this limitation, we adopt the Fine-grained Adversarial Domain Adaptation (FADA) framework proposed in \cite{fada}, as illustrated in Figure~\ref{fig:domain_adaptation_architecture}b. The key innovation of FADA is the introduction of a fine-grained domain discriminator that operates on a per-pixel basis. This discriminator distinguishes between source and target features but also considers the semantic class (e.g., blood vessel versus background) associated with each pixel.

In practice, the segmentation network produces softmax probability maps for each class on both source and target images. For the source domain, these probability maps are compared against the ground truth labels. For the target domain, where annotations are absent, the network’s own predictions serve as pseudo-labels. By enforcing adversarial learning at the level of individual classes, FADA promotes a more precise alignment of feature distributions across domains.



\subsection{Dimensionality Reduction} 
Another challenge arises from the discrepancy in channel dimensionality between the source and target datasets. The segmentation model is pre-trained on grayscale or RGB images (1 or 3 channels), whereas the hyperspectral data comprises 826 channels. Consequently, we need to reduce the hyperspectral data's dimensionality to align with the channel size of the FIVES dataset. We explore two complementary strategies:

\subsubsection*{Static Dimensionality Reduction}
In our static approach, we apply median windowing across specific wavelength ranges to condense the hyperspectral data to a single channel. We focus primarily on the 500–600 nm range, corresponding to the peak absorption of hemoglobin \cite{hemoglobinAbsorption}. Since hemoglobin molecules, which are proteins found in red blood cells \cite{hemoglobinDef}, absorb strongly in this range, the resulting windowed image exhibits high contrast between blood vessels and surrounding soft tissues.

\subsubsection*{Learnable Dimensionality Reduction}
\begin{figure}[t]
  \centering
  \includegraphics[width=\columnwidth]{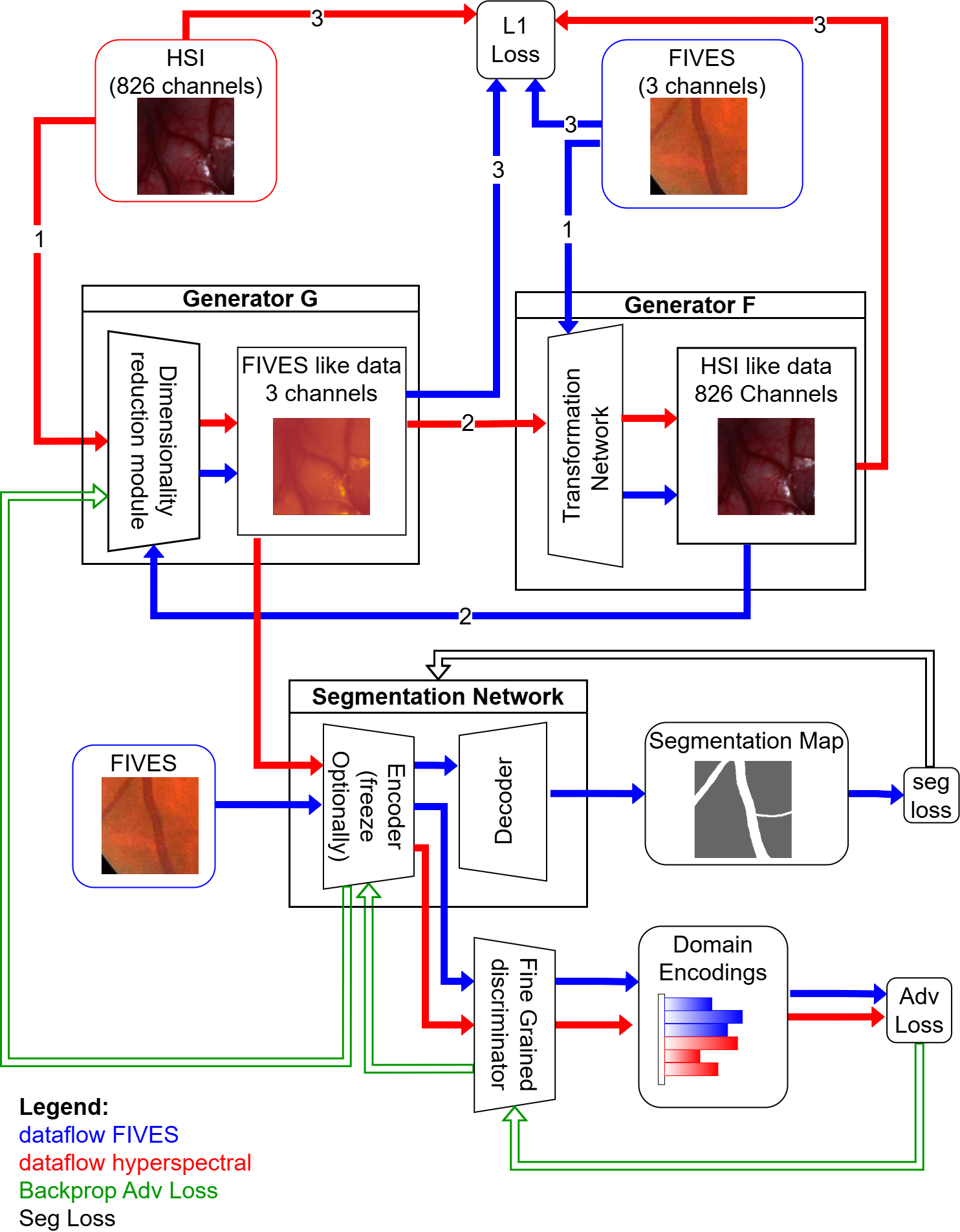}
  \caption{FADA architecture with learnable dimensionality reduction based on cycle loss.}
  \label{fig:FADA_CycleGAN}
\end{figure}

Drawing inspiration from CycleGAN \cite{cycleGAN}, we incorporate a learnable dimensionality reduction strategy into our architecture, as depicted in Figure~\ref{fig:FADA_CycleGAN}. In this framework, a CNN-based Generator \(G\) is trained to map the hyperspectral data to a reduced representation (either a single or an RGB channel). A complementary Generator \(F\), serving as a decoder, reconstructs the hyperspectral image from the reduced representation. Additionally, we transform the source data (FIVES) to the hyperspectral domain using \(F\) and then reconstruct it using \(G\). Both generators are optimized using an L1 reconstruction loss. Furthermore, the fine-grained discriminator from the FADA architecture is employed to impose an adversarial domain loss.

\section{Experiments}
\subsection{Set-up}
\begin{figure}[!h]
  \centering
  \includegraphics[height=0.53\textheight,keepaspectratio]{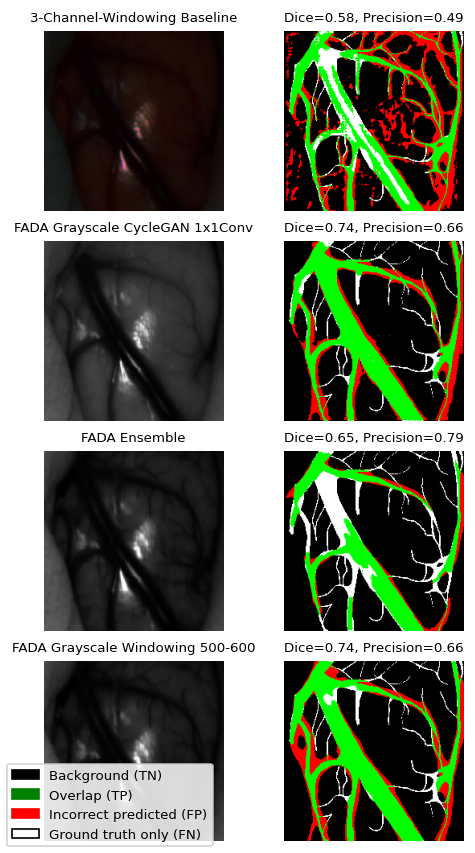}
  \caption{Domain adaptation results, with dimensionally reduced hyperspectral images on the left and the corresponding predicted segmentation masks on the right.}
  \label{fig:segmentation_prediction}
\end{figure}

HELICoiD dataset comprises 36 images with sparse annotations. We enlisted domain experts to complete the sparse vessel annotations for 5 of them (see Fig. \ref{fig:segmentation_prediction}). For training purposes, we use the remaining 31 samples. We employ a cross-testing strategy for both hyperparameter tuning and evaluation. We train 10 models by randomly sampling from the hyperparameter space and then select the best-performing model, measured by the Dice score, using one of the annotated samples. The model corresponding to the best Dice score is subsequently tested on the remaining 4 labeled images. This process is repeated five times, each time using a different validation sample and test subset.

In total, we evaluated 9 distinct domain adaptation approaches. For all approaches, we use the LinkNet \cite{linknet} architecture combined with a RegNet \cite{regnetx} encoder:

\begin{enumerate}
    \item \textbf{FADA-Grayscale-Windowing-500to600}: Static dimensionality reduction to a single channel via median windowing in the 500–600 nm range.
    \item \textbf{FADA-Grayscale-Windowing-500to600 (No Pre-training)}: Similar to the previous approach, the segmentation model is trained exclusively within the FADA pipeline without pre-training on the FIVES dataset.
    \item \textbf{FADA-3Channel-Windowing}: Static dimensionality reduction to RGB channels using median windowing in the ranges 600–1000 nm, 500–600 nm, and 400–500 nm.
    \item \textbf{FADA-Grayscale-Windowing-400to500}: Static dimensionality reduction to a single channel using median windowing in the 400–500 nm range.
    \item \textbf{FADA-Grayscale-Windowing-600to800}: Static dimensionality reduction to a single channel using median windowing in the 600–800 nm range.
    \item \textbf{FADA-Ensemble}: An ensemble model that averages the predictions of the FADA-Grayscale-Windowing-400to500, FADA-Grayscale-Windowing-500to600, and FADA-Grayscale-Windowing-600to800 approaches.
    \item \textbf{FADA-Grayscale-Windowing-500to600-CLDice}: A variant of the FADA-Grayscale-Windowing-500to600 approach that employs a Centerline Dice loss \cite{cldice} for segmentation.
    \item \textbf{FADA-Grayscale-CycleGAN-CNN}: A learnable dimensionality reduction approach that employs a CycleGAN framework to reduce the hyperspectral data to grayscale using a CNN.
    \item \textbf{FADA-Grayscale-CycleGAN-1x1conv}: A learnable dimensionality reduction approach using the CycleGAN framework, but with a simplified channel reduction achieved solely via a 1x1 convolutional layer, without a CNN.
\end{enumerate}

\begin{table*}[t]
\centering
\caption{Average Metric Comparison Across Approaches}
\label{tab:metrics}
\resizebox{\textwidth}{!}{%
\begin{tabular}{@{}lccccc@{}}
\toprule
\textbf{Approach} & \textbf{Precision} & \textbf{Recall} & \textbf{Dice Score} & \textbf{Accuracy} & \textbf{CLDice Score} \\ \midrule
Baseline Grayscale                         & 0.64 $\pm$ 0.03 & 0.27 $\pm$ 0.05 & 0.33 $\pm$ 0.05 & 0.82 $\pm$ 0.01 & 0.30 $\pm$ 0.04 \\
Baseline 3-Channel-Windowing               & 0.40 $\pm$ 0.04 & 0.51 $\pm$ 0.03 & 0.45 $\pm$ 0.04 & 0.75 $\pm$ 0.01 & 0.47 $\pm$ 0.03 \\\midrule
Approach 1: FADA-Grayscale-Windowing-500to600   & 0.53 $\pm$ 0.01 & \textbf{0.66 $\pm$ 0.06} & 0.58 $\pm$ 0.03 & 0.82 $\pm$ 0.02 & \textbf{0.54 $\pm$ 0.03} \\
Approach 2: FADA-Grayscale-Windowing-500to600-NoPretraining & 0.49 $\pm$ 0.05 & 0.54 $\pm$ 0.10 & 0.50 $\pm$ 0.04 & 0.79 $\pm$ 0.01 & 0.45 $\pm$ 0.06 \\
Approach 3: FADA-3Channel-Windowing            & 0.50 $\pm$ 0.03 & 0.63 $\pm$ 0.06 & 0.55 $\pm$ 0.04 & 0.80 $\pm$ 0.01 & 0.53 $\pm$ 0.04 \\
Approach 4: FADA-Grayscale-Windowing-400to500   & 0.55 $\pm$ 0.06 & 0.59 $\pm$ 0.12 & 0.55 $\pm$ 0.06 & 0.82 $\pm$ 0.01 & 0.47 $\pm$ 0.05 \\
Approach 5: FADA-Grayscale-Windowing-600to800   & 0.61 $\pm$ 0.04 & 0.60 $\pm$ 0.08 & 0.59 $\pm$ 0.04 & \textbf{0.84 $\pm$ 0.01} & 0.47 $\pm$ 0.05 \\
Approach 6: FADA-Ensemble                       & \textbf{0.69 $\pm$ 0.07} & 0.41 $\pm$ 0.03 & 0.50 $\pm$ 0.04 & \textbf{0.84 $\pm$ 0.01} & 0.38 $\pm$ 0.02 \\
Approach 7: FADA-Grayscale-Windowing-500to600-CLDice & 0.40 $\pm$ 0.20 & 0.61 $\pm$ 0.16 & 0.50 $\pm$ 0.10 & 0.78 $\pm$ 0.02 & 0.45 $\pm$ 0.23 \\
Approach 8: FADA-Grayscale-CycleGAN-CNN       & 0.60 $\pm$ 0.05 & 0.57 $\pm$ 0.09 & 0.57 $\pm$ 0.05 & 0.83 $\pm$ 0.01 & 0.44 $\pm$ 0.05 \\
Approach 9: FADA-Grayscale-CycleGAN-1x1Conv     & 0.58 $\pm$ 0.02 & 0.63 $\pm$ 0.03 & \textbf{0.60 $\pm$ 0.02} & \textbf{0.84 $\pm$ 0.01} & 0.48 $\pm$ 0.01 \\
\bottomrule
\end{tabular}%
}
\end{table*}

\textit{Baseline}. For comparison, we considered two baseline models: one trained solely on the FIVES dataset with static median windowing in the 500–600 nm range and the other using RGB windowing (600–1000 nm, 500–600 nm, and 400–500 nm) as hyperspectral dimensionality reduction. We also attempted to train a U-Net directly on the hyperspectral data using the sparsely annotated blood-vessel labels from the HELICoiD dataset. However, the model defaulted to predicting false negatives due to the scarcity of annotations.

\subsection{Results}
The results, summarized in Table 1, indicate that all FADA-based approaches significantly outperform the baseline models. Notably, the FADA-Grayscale-CycleGAN-1x1Conv method yields the best performance with an average dice score of 0.60, outperforming both static windowing methods and other variants of the FADA approach. This improvement demonstrates the benefit of employing a learnable dimensionality reduction strategy within the FADA framework, as it enhances the model's ability to adapt to the hyperspectral domain.

\section{Conclusion} 
In this work, we investigated various domain adaptation strategies for blood vessel segmentation in hyperspectral brain images. By leveraging both static and learnable dimensionality reduction techniques within a fine-grained adversarial framework, our proposed FADA-based methods substantially outperformed traditional baseline models. 

\section*{Acknowledgment}
This work was supported by the European Union’s Horizon Research programme (Grant No. 101071040).

\ifCLASSOPTIONcaptionsoff
  \newpage
\fi

\bibliography{example_paper}

\begin{thebibliography}{10}

\bibitem{HSIclinical}
A.~N. Sen, S.~P. Gopinath, and C.~S. Robertson, ``Clinical application of near-infrared spectroscopy in patients with traumatic brain injury: a review of the progress of the field,'' {\em Neurophotonics}, vol.~3, no.~3, pp.~031409--031409, 2016.

\bibitem{unet}
O.~Ronneberger, P.~Fischer, and T.~Brox, ``U-net: Convolutional networks for biomedical image segmentation,'' in {\em Medical Image Computing and Computer-Assisted Intervention -- MICCAI 2015} (N.~Navab, J.~Hornegger, W.~M. Wells, and A.~F. Frangi, eds.), (Cham), pp.~234--241, Springer International Publishing, 2015.

\bibitem{fives}
K.~Jin, X.~Huang, J.~Zhou, Y.~Li, Y.~Yan, Y.~Sun, Q.~Zhang, Y.~Wang, and J.~Ye, ``Fives: A fundus image dataset for artificial intelligence based vessel segmentation,'' {\em Scientific data}, vol.~9, no.~1, p.~475, 2022.

\bibitem{helicoid}
H.~Fabelo, S.~Ortega, A.~Szolna, D.~Bulters, J.~F. Piñeiro, S.~Kabwama, A.~J-O'Shanahan, H.~Bulstrode, S.~Bisshopp, B.~R. Kiran, D.~Ravi, R.~Lazcano, D.~Madroñal, C.~Sosa, C.~Espino, M.~Marquez, M.~De~La Luz~Plaza, R.~Camacho, D.~Carrera, M.~Hernández, G.~M. Callicó, J.~Morera~Molina, B.~Stanciulescu, G.-Z. Yang, R.~Salvador, E.~Juárez, C.~Sanz, and R.~Sarmiento, ``In-vivo hyperspectral human brain image database for brain cancer detection,'' {\em IEEE Access}, vol.~7, pp.~39098--39116, 2019.

\bibitem{DA_GRL}
Y.~Ganin and V.~Lempitsky, ``Unsupervised domain adaptation by backpropagation,'' in {\em International conference on machine learning}, pp.~1180--1189, PMLR, 2015.

\bibitem{fada}
H.~Wang, T.~Shen, W.~Zhang, L.-Y. Duan, and T.~Mei, ``Classes matter: A fine-grained adversarial approach to cross-domain semantic segmentation,'' in {\em European conference on computer vision}, pp.~642--659, Springer, 2020.

\bibitem{hemoglobinAbsorption}
W.~Zijlstra and A.~Buursma, ``Spectrophotometry of hemoglobin: absorption spectra of bovine oxyhemoglobin, deoxyhemoglobin, carboxyhemoglobin, and methemoglobin,'' {\em Comparative Biochemistry and Physiology Part B: Biochemistry and Molecular Biology}, vol.~118, no.~4, pp.~743--749, 1997.

\bibitem{hemoglobinDef}
A.~N. Schechter, ``Hemoglobin research and the origins of molecular medicine,'' {\em Blood, The Journal of the American Society of Hematology}, vol.~112, no.~10, pp.~3927--3938, 2008.

\bibitem{cycleGAN}
J.-Y. Zhu, T.~Park, P.~Isola, and A.~A. Efros, ``Unpaired image-to-image translation using cycle-consistent adversarial networks,'' in {\em Proceedings of the IEEE international conference on computer vision}, pp.~2223--2232, 2017.

\bibitem{linknet}
A.~Chaurasia and E.~Culurciello, ``Linknet: Exploiting encoder representations for efficient semantic segmentation,'' in {\em 2017 IEEE visual communications and image processing (VCIP)}, pp.~1--4, IEEE, 2017.

\bibitem{regnetx}
I.~Radosavovic, R.~P. Kosaraju, R.~Girshick, K.~He, and P.~Doll{\'a}r, ``Designing network design spaces,'' in {\em Proceedings of the IEEE/CVF conference on computer vision and pattern recognition}, pp.~10428--10436, 2020.

\bibitem{cldice}
S.~Shit, J.~C. Paetzold, A.~Sekuboyina, I.~Ezhov, A.~Unger, A.~Zhylka, J.~P. Pluim, U.~Bauer, and B.~H. Menze, ``cldice-a novel topology-preserving loss function for tubular structure segmentation,'' in {\em Proceedings of the IEEE/CVF conference on computer vision and pattern recognition}, pp.~16560--16569, 2021.

\end{thebibliography}
\bibliographystyle{ieeetr}

\end{document}